\documentclass[final]{cvpr}
\usepackage{times}
\usepackage{epsfig}
\usepackage{graphicx}
\usepackage{amsmath}
\usepackage{amssymb}
\usepackage{enumitem}
\usepackage{bbm}
\usepackage{multirow}
\usepackage{booktabs}
\usepackage[table]{xcolor}

\definecolor{citecolor}{HTML}{0071bc}
\usepackage[pagebackref=false,breaklinks=true,letterpaper=true,colorlinks,citecolor=citecolor,bookmarks=false]{hyperref}
\usepackage{complexity}
\usepackage{subcaption}
\usepackage{thm-restate}
\usepackage{amsthm}
\usepackage{amsmath}
\usepackage{mathtools}
\usepackage{tabularx}
\usepackage{fancyvrb}
\usepackage{comment} 
\usepackage{algorithm}
\usepackage{algorithmic}

\newcommand{\domain}{\mathcal{D}}

\newcommand{\data}{\mathcal{X}}
\newcommand{\hdata}{\mathcal{Z}}
\newcommand{\ydata}{\mathcal{Y}}

\newcommand{\err}{\varepsilon}
\newcommand{\emperr}{\widehat{\varepsilon}}
\newcommand{\RR}{\mathbb{R}}

\newcommand{\Exp}{\mathbb{E}}
\newcommand{\HH}{\mathcal{H}}
\newcommand{\xx}{\mathbf{x}}

\newcommand{\defeq}{\vcentcolon=}

\newcommand{\eps}{\varepsilon}

\newcommand{\dist}{\mathcal{D}}

\theoremstyle{definition}
\newtheorem{definition}{Definition}[section]

\newtheorem{theorem}{Theorem}[section]
\newtheorem{remark}{Remark}[section]

\newtheorem{lemma}{Lemma}[section]
\newtheorem{proposition}{Proposition}[section]

\DeclarePairedDelimiter\set\{\}

\def\cvprPaperID{11397} 
\def\confYear{CVPR 2021}
\begin{document}
\title{Learning Invariant Representations and Risks \\ for Semi-supervised Domain Adaptation}


\author{Bo Li$^{13}$\thanks{Equal contribution.}
\quad
Yezhen Wang$^{23}$\footnotemark[1]
\quad
Shanghang Zhang$^{1}$\footnotemark[1]
\quad
\quad
Dongsheng Li$^3$
\\
Kurt Keutzer$^1$
\quad
Trevor Darrell$^1$
\quad
Han Zhao$^4$\thanks{Work done while at Carnegie Mellon University}
\\
$^1$BAIR, UC Berkeley
\quad
$^2$UC San Diego
\quad
$^3$Microsoft Research Asia
\quad $^4$UIUC
}

\maketitle

\begin{abstract}
The success of supervised learning hinges on the assumption that the training and test data come from the same underlying distribution, which is often not valid in practice due to potential distribution shift.
In light of this, most existing methods for unsupervised domain adaptation focus on achieving domain-invariant representations and small source domain error. However, recent works have shown that this is not sufficient to guarantee good generalization on the target domain, and in fact, is provably detrimental under label distribution shift. Furthermore, in many real-world applications it is often feasible to obtain a small amount of labeled data from the target domain and use them to facilitate model training with source data. Inspired by the above observations, in this paper we propose the first method that aims to simultaneously learn invariant representations and risks under the setting of semi-supervised domain adaptation (Semi-DA). First, we provide a finite sample bound for both classification and regression problems under Semi-DA. The bound suggests a principled way to obtain target generalization, i.e., by aligning both the marginal and conditional distributions across domains in feature space. 
Motivated by this, we then introduce the LIRR algorithm for jointly Learning Invariant Representations and Risks. Finally, extensive experiments are conducted on both classification and regression tasks, which demonstrate that LIRR consistently achieves state-of-the-art performance and significant improvements compared with the methods that only learn invariant representations or invariant risks. Our code will be released at \href{https://github.com/Luodian/Learning-Invariant-Representations-and-Risks}{LIRR@github}
\end{abstract}

\vspace{-0.2cm}
\section{Introduction}
The success of supervised learning hinges on the key assumption that test data should share the same distribution with the training data. Unfortunately, in most of the real-world applications, data are dynamic, meaning that there is often a distribution shift between the training (source) and test (target) domains. To this end, unsupervised domain adaptation (UDA) methods aim to approach this problem by adapting the predictive model from labeled source data to the unlabeled target data. Recent advances in UDA focus on learning domain-invariant representations that also lead to a small error on the source domain. The goal is to learn representations, along with the source predictor, that can generalize to the target domain~\cite{long2015learning, ganin2016domain, tzeng2017adversarial, long2018conditional, chen2019homm,zhao2018adversarial}. However, recent works~\cite{zhao2019learning, wu2019domain, combes2020domain} have shown that the above conditions are not sufficient to guarantee good generalizations on the target domain. In fact, if the marginal label distributions are distinct across domains, the above method provably hurts target generalization~\cite{zhao2019learning}.

On the other hand, while labeled target data is usually more difficult or costly to obtain than labeled source data, it can lead to better accuracy~\cite{hanneke2019value}. Furthermore, in many practical applications, e.g., vehicle counting, object detection, speech recognition, etc., it is often feasible to at least obtain a small amount of labeled data from the target domain so that it can facilitate model training with source data~\cite{li2018semi,saito2019semi}. Motivated by these observations, in this paper we focus on a more realistic setting of semi-supervised domain adaptation (Semi-DA). In Semi-DA, in addition to the large amount of labeled source data, the learner also has access to a small amount of labeled data from the target domain. Again, the learner's goal is to produce a hypothesis that well generalizes to the target domain, under the potential shift between the source and the target. Semi-DA is a more-realistic setting that allows practitioners to design better algorithms that can overcome the aforementioned limitations in UDA. The key question in this scenario is: \emph{how to maximally exploit the labeled target data for better model generalization?} 

\begin{figure*}[!tb]
    \centering
    \includegraphics[width=1\linewidth]{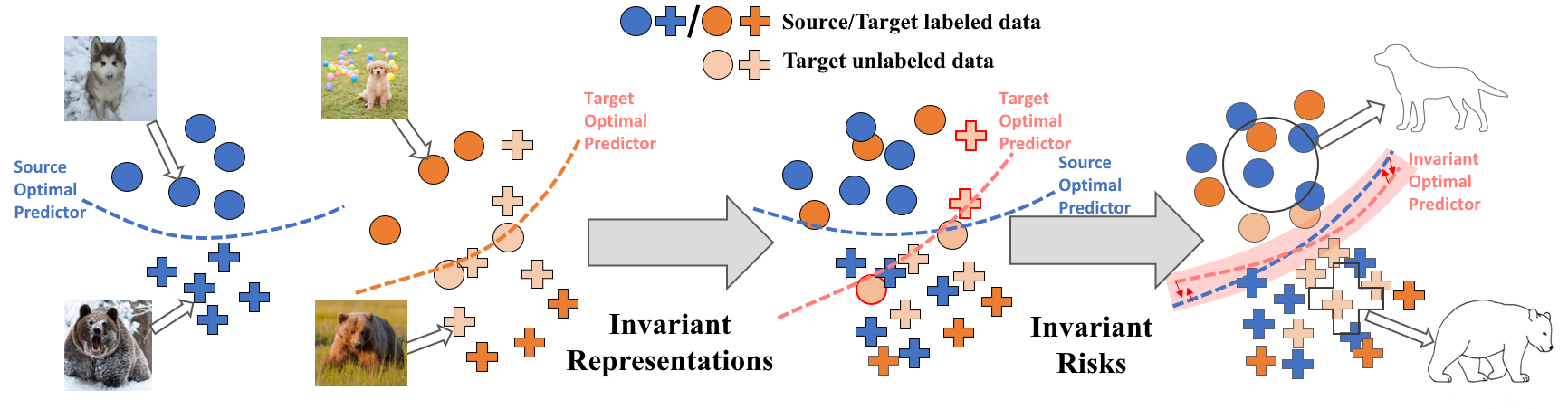}
    \caption{Overview of the proposed model. Learning invariant representations induces indistinguishable representations across domains, but there can still be mis-classified samples (as stated in red circle) due to misaligned optimal predictors. Besides learning invariant representations, LIRR model jointly learns invariant risks to better align the optimal predictors across domains.}
    \label{fig:problem}
\end{figure*}

In this paper, we address the above question under the Semi-DA setting. In order to first understand how performance discrepancy occurs, we derive a finite-sample generalization bound for both classification and regression problems under Semi-DA. Our theory shows that, for a given predictor, the accuracy discrepancy between two domains depends on two terms: (i) the distance between the marginal feature distributions, and (ii) the distance between the optimal predictors from source and target domains. Our observation naturally leads to a principled way of learning invariant representations (to minimize discrepancy between marginal feature distributions) and risks (to minimize discrepancy between conditional distributions over the features) across domains simultaneously for a better generalization on the target. In light of this, we introduce our novel bound minimization algorithm LIRR, a model of jointly Learning Invariant Representations and Risks for such purposes. As a comparison, existing works either focus on learning invariant representations only~\cite{ganin2016domain, tzeng2017adversarial,zhao2018adversarial,chen2019homm}, or learning invariant risks only~\cite{arjovsky2019invariant, chang2020invariant, song2019learning, steinberg2020fast, krueger2020out, zhao2020domain}, but not both. However, these are not sufficient to reduce the accuracy discrepancy for good generalizations on the target. To our best knowledge, LIRR is the first work that subtly combine above learning objectives with sound theoretical justification. LIRR jointly learns invariant representations and risks, and as a result, better mitigates the accuracy discrepancy across domains. To better understand our method, we illustrate the proposed algorithm, LIRR, in Fig.~\ref{fig:problem}.

In summary, our work provides the following contributions:
\begin{itemize}[leftmargin=*]
\setlength\itemsep{0em}
     \item Theoretically, we provide finite-sample generalization bounds for Semi-DA on both classification (Theorem~\ref{eq:classification_upper_bound}) and regression (Theorem~\ref{eq:regression_upper_bound}) problems. Our bounds inform new directions for simultaneously optimizing both marginal and conditional distributions across domains for better generalization on the target. To the best of our knowledge, this is the first generalization analysis in the Semi-DA setting that takes into account both the shifts between the marginal and the conditional distributions from source and target domains.  

     \item To bridge the gap between theory and practice, we provide an information-theoretic interpretation of our theoretical results. Based on this perspective, we propose a bound minimization algorithm, LIRR, to jointly learn invariant representations and invariant optimal predictors, in order to mitigate the accuracy discrepancy across domains for better generalizations. 
     
     \item We systematically analyze LIRR with extensive experiments on both classification and regression tasks. Compared with methods that only learn invariant representations or invariant risks, LIRR demonstrates significant improvements on Semi-DA. We also analyze the adaptation performance with an increasing amount of labeled target data, which shows LIRR even surpasses oracle method \emph{Full Target} trained only on labeled target data, suggesting that LIRR can successfully exploit the structure in source data to improve generalization on the target domain. 
\end{itemize}

\section{Related Work}
\subsection{Domain Adaptation} 
Most existing research on domain adaptation focuses on the unsupervised setting, \textit{i.e.} the data from target domain are fully unlabeled. Recent deep unsupervised domain adaptation (UDA) methods usually employ a conjoined architecture with two streams to represent the models for the source and target domains, respectively~\cite{zhuo2017deep}. Besides the task loss on the labeled source domain, another alignment loss is designed to align the source and target domains, such as discrepancy loss~\cite{long2015learning,sun2016return,zhuo2017deep,adel2017unsupervised,kang2019contrastive,chen2020homm}, adversarial loss~\cite{bousmalis2017unsupervised,tzeng2017adversarial,shrivastava2017learning,russo2018source,zhao2019multi}, and self-supervision loss~\cite{ghifary2015domain,ghifary2016deep,bousmalis2016domain,carlucci2019domain,feng2019self,kim2020cross,mei2020instance}.
Semi-DA deals with the domain adaptation problem where some target labels are available~\cite{donahue2013semi,li2014learning,yao2015semi,ao2017fast}. \cite{saito2019semi} empirically observed that UDA methods often fail in improving accuracy in Semi-DA and proposed a min-max entropy approach that adversarially optimizes an adaptive few-shot model. Different from these works, our proposed method aims to align \emph{both} the marginal feature distributions as well as the conditional distributions of the label over the features, which can overcome the limitations that exist in UDA methods that only align feature distributions~\cite{zhao2019learning}.

\subsection{Invariant Risk Minimization}
In a seminal work, \cite{arjovsky2019invariant} consider the question that data are collected from multiple envrionments with different distributions where spurious correlations are due to dataset biases. This part of spurious correlation will confuse model to build predictions on unrelated correlations \cite{lake2017building, janzing2010causal, scholkopf2012causal} rather than true causal relations. IRM~\cite{arjovsky2019invariant} estimates invariant and causal variables from multiple environments by regularizing on predictors to find data represenation matching for all environments. \cite{chang2020invariant} extends IRM to neural predictions and employ the environment aware predictor to learn a rationale feature encoder. As a comparison, in this work we argue that IRM is not sufficient to ensure reduced accuracy discrepancy across domains, and we propose to align the marginal features as well simultaneously.

\section{Preliminaries}


\subsection{Unsupervised Domain Adaptation}
We use $\data$ and $\ydata$ to denote the input and output space, respectively. Similarly, $\hdata$ stands for the representation space induced from $\data$ by a feature transformation $g:\data\mapsto\hdata$. Accordingly, we use $X, Y, Z$ to denote random variables which take values in $\data, \ydata,\hdata$. Throughout the paper, a domain corresponds to a joint distribution on the input space $\data$ and output space $\ydata$. We use $\dist_S$ ($\dist_T$) to denote the source (target) domain and subsequently we also use $\dist_S(Z) (\dist_T(Z))$ to denote the marginal distributions of $\dist_S (\dist_T)$ over $Z$. Furthermore, let $D$ be a categorical variable that corresponds to the index of domain, i.e., $D\in\{S, T\}$. The overall sampling process for our data can then be specified by first drawing a value of $D$, and then depending on the value of $D$, we sample from the corresponding distribution $\dist_D$. Under this setting, the probabilities of $\Pr(D = T)$ and $\Pr(D = S)$ then determine the relative sample sizes of our target and source data.  

A hypothesis over the feature space $\hdata$ is a function $h:\hdata\to [0, 1]$. The \emph{error} of a hypothesis $h$ under distribution $\dist_S$ and feature transformation $g$ is defined as: $\eps_S(h, f)\defeq \Exp_{\dist_S}[|h(g(X)) - f(X)|]$. In classification setting, in which $f$ and $h$ are binary classification functions, above definition reduces to the probability that $h$ disagrees with $f$ under $\mathcal{D}_S : \Exp_{\dist_S}[|h(g(X)) - f(X)|] = \Pr_{\dist_S}(h(g(X))\neq Y)$. In regression, the above error is then the usual mean absolute error, i.e., the $\ell_1$ loss. As a common notation, we also use $\widehat{\eps}_S(h)$ to denote the empirical risk of $h$ on the source domain. Similarly, $\eps_T(h)$ and $\widehat{\eps}_T(h)$ are the true risk and the empirical risk on the target domain. For a hypothesis class $\HH$, we use $VCdim(\HH)$ and $Pdim(\HH)$ to denote the VC-dimension and pseudo-dimension of $\HH$, respectively.

\subsection{Semi-supervised Domain Adaptation}
Formally, in Semi-DA the learner is allowed to have access to a small amount of labeled data in target domain $\dist_T$. Let $S = \{(\xx^{(S)}_i, y^{(S)}_i)\}_{i=1}^n$ be a set of labeled data sampled i.i.d.\ from $\dist_S$. Similarly, we have $T = \{(\xx^{(T)}_j)\}_{j=1}^k$ as the set of target unlabeled data sampled from $\dist_T$, and we let $\tilde{T} = \{(\xx^{(\tilde{T})}_j, y^{(\tilde{T})}_j)\}_{j=1}^m$ be the small set of labeled data where $m \leq k$. Usually, we also have $m \ll n$, and the goal of the learner is to find a hypothesis $h\in\HH$ by learning from $S$, $T$ and $\tilde{T}$ so that $h$ has a small target error $\err_T(h)$. 


Clearly, with the additional small amount of labeled data $\tilde{T}$, one should expect a better generalization performance than what the learner could hope to achieve in the setting of unsupervised domain adaptation. To this end, we first state the following generalization upper bound from~\cite{zhao2019learning} in the setting of unsupervised domain adaptation:
\begin{theorem}\cite{zhao2019learning}
Let $\langle \domain_S(X), f_S\rangle$ and $\langle \domain_T(X), f_T\rangle$ be the source and target domains. For any function class $\HH\subseteq [0,1]^{\data}$, and $\forall h\in\HH$, the following inequality holds: 
\begin{equation}
\begin{aligned}
\err_T(h) \leq &~ \err_S(h)  + d_{\HH}(\domain_S(X), \domain_T(X)) \\
& + \min\{\Exp_{\domain_S}[|f_S - f_T|], \Exp_{\domain_T}[|f_S - f_T|]\}.
\end{aligned}
\end{equation}
\label{thm:populationbound}
\end{theorem}
The $d_{\HH}(\cdot, \cdot)$ is known as the $\HH$-divergence~\cite{ben2010theory}, a pseudo-metric parametrized by $\HH$ to measure the discrepancy between two distributions. It should be noted that the above theorem is a \emph{population result}, hence it does not give a \emph{finite sample bound}. Furthermore, the setting above is \emph{noiseless}, where $f_S$ and $f_T$ correspond to the groundtruth labeling functions in source and target domains. Nevertheless, it provides an insight on achieving domain adaptation through bounding the error difference on source and target domains: to simultaneously minimize the distances between feature representations and between the optimal labeling functions. In the next section we shall build on this result to derive finite sample bound in semi-supervised domain adaptation.

\section{Generalization Bounds for Semi-supervised Domain Adaptation}
\label{sec:bound}
In this section, we derive a finite-sample generalization bound for Semi-DA, where the model has access to both a large amount of labeled data $S$ from the source domain, and a small amount of labeled data $\tilde{T}$ from the target domain. 
For this purpose, we first introduce the definition of $\mathcal{H}$ on both classification and regression settings, and then present our theoretical results of the generalization upper bounds for Semi-DA. 

\begin{definition}
Let $\HH$ be a family of binary functions from $\hdata$ to $\{0, 1\}$, and $\mathcal{A}_{\mathcal{H}}$ be the collection of subsets of $\hdata$ defined as $\mathcal{A}_{\mathcal{H}} \defeq \{h^{-1}(1) \ | \ h \in \HH \}$. The distance between two distributions $\mathcal{D}$ and $\mathcal{D}'$ based on $\mathcal{H}$ is: $d_{\mathcal{H}} (\dist, \dist') \defeq  \text{sup}_{A \in \mathcal{A}_{\HH}} |\Pr_{\dist} (A) - \Pr_{\dist'}(A)|$. 
\end{definition}
With the above definition, we have the symmetric difference w.r.t.\ itself as: $\HH\Delta\HH = \{h(z) \oplus h'(z)\ | \ h, h' \in \HH\}$, where $\oplus$ is the \textbf{XOR} operation. Next, considering that for a joint distribution $\dist$ over $\hdata\times\ydata$ in our setting, there may be noise in the conditional distribution $\Pr_\dist(Y \mid Z)$. It is then necessary to define a term to measure the noise level of each domain. To this end, in classification, we define the noise on the source domain $n_S \defeq \Exp_{S}[|Y - f_S(Z)|]$, where $f_S:\hdata\to[0, 1]$ is the conditional mean function, i.e., $f_S(Z) = \Exp_{S}[Y \mid Z]$. Similar definition also applies to the target domain, where we use $n_T$ to denote the noise in target. In regression, with $\ell_1$ loss, we define $f_S: \hdata \to \RR$ to be the conditional median function of $\Pr(Y \mid Z)$, i.e. $f_S(Z) := \inf_y \{y \in \mathbb{R}: 1/2 \leq \Pr(Y \leq y \mid Z)\}$. Now we are ready to state the main results in this section:

\begin{restatable}{theorem}{clsupperbound}
\vspace*{-1em}
 (Classification generalization bound in Semi-DA). Let $\HH$ be a hypothesis set with functions $h: \hdata \to \{0,1\}$ and $VCdim(\HH) = d$, $\widehat{\mathcal{D}}_S$ (resp. $\widehat{\mathcal{D}}_T$) be the empirical distribution induced by samples from $\dist_S$ (resp. $\dist_T$). For $0 < \delta < 1$, then w.p.\ at least $1 - \delta$ over the $n$ samples in $S$ and $m$ samples in $\tilde{T}$, for all $h \in \HH$, we have:
\label{eq:classification_upper_bound}
\begin{align*}
    ~& \eps_T(h) \leq \frac{m}{n+m}\emperr_{\tilde{T}}(h) + \frac{n}{n + m}\emperr_S(h)  \\
    & + \frac{n}{n + m} \set[\Big]{ d_{\HH\Delta\HH} (\widehat{\dist}_S(Z), \widehat{\dist}_T(Z)) \  + \\
    & \quad \min\{\Exp_S[|f_S(Z) - f_{\tilde{T}}(Z)|], \Exp_T[|f_S(Z) - f_{\tilde{T}}(Z)|]\}} \\
    & + \frac{n}{n+m} |n_S +n_{\tilde{T}}| \\
    & + O \left( \sqrt{(\frac{1}{m} + \frac{1}{n}) \text{log} \frac{1}{\delta} + \frac{d}{n} \text{log}\frac{n}{d} + \frac{d}{m} \text{log}\frac{m}{d}} \right).
\end{align*}
\end{restatable}

Next, by replacing the VC dimension in the above theorem with Pseudo-dimension ($Pdim$), we can also prove a corresponding generalization bound in regression as well:
\begin{restatable}{theorem}{regrupperbound}
\vspace*{-1em}
(Regression generalization bound in Semi-DA). Let $\HH$ be a hypothesis set with functions $h: \hdata \to [0, 1]$ and $Pdim(\HH) = d$, $\widehat{\mathcal{D}}_S$ (resp. $\widehat{\mathcal{D}}_T$) be the empirical distribution induced by samples from $\dist_S$ (resp. $\dist_T$). Then we define $\tilde{\HH} := \{\mathbb{I}_{|h(x) - h'(x)| > t} : h, h' \in \HH, 0 \leq t \leq 1 \}$. For $0 < \delta < 1$, then w.p.\ at least $1 - \delta$ over the $n$ samples in $S$ and $m$ samples in $\tilde{T}$, for all $h \in \HH$, we have:
 \label{eq:regression_upper_bound}
\begin{align*}
    ~&\eps_T(h) \leq \frac{m}{n+m}\emperr_{\tilde{T}}(h) + \frac{n}{n + m}\emperr_S(h) \\ 
    & + \frac{n}{n + m} \set[\Big]{ d_{\tilde{\HH}} (\widehat{\dist}_S(Z), \widehat{\dist}_T(Z)) \ +  \\
    & \quad \min\{\Exp_S[|f_S(Z) - f_{\tilde{T}}(Z)|], \Exp_T[|f_S(Z) - f_{\tilde{T}}(Z)|]\}} \\
    & + \frac{n}{n+m} |n_S + n_{\tilde{T}}| \\
    & + O \left( \sqrt{(\frac{1}{m} + \frac{1}{n}) \text{log} \frac{1}{\delta} + \frac{d}{n} \text{log}\frac{n}{d} + \frac{d}{m} \text{log}\frac{m}{d}} \right).
\end{align*}
\end{restatable}



\begin{remark}
It is worth pointing out that both $n_S$ and $n_T$ are constants that only depend on the underlying source and target domains, respectively. Hence $|n_S +n_T|$ essentially captures the the amplitude of noise. The last two terms of the bound come from standard concentration analysis for uniform convergence. 

Due to space limit, we leave more discussions on how to extend from binary to multi-class, from $\ell_1$ to $\ell_p$ loss, and bridging the gap between theory and practice in appendix.
\end{remark}

Compared with previous results~\cite{ben2010theory,zhao2019learning, combes2020domain, redko2017theoretical, redko2019advances}, our bounds is the \emph{first} in the Semi-DA literature which contains empirical error terms from \emph{both} the source and target domains and free of the joint optimal errors term, e.g., the $\lambda$ in Theorem 3 of~\cite{ben2010theory}. The difference here is significant since the joint optimal errors depend on the choice of the hypothesis class $\HH$ and in fact it can change arbitrarily as the feature space changes. In fact, it has been recently shown that the change of $\lambda$ during representation learning is precisely the cause that fails classic domain invariant learning in the setting of unsupervised domain adaptation. Furthermore, these bounds imply a natural and principled way for a better generalization to the target domain by learning invariant representations and risks simultaneously. Note that this is in sharp contrast to previous works where only invariant representations are pursued~\cite{ganin2016domain,zhao2018adversarial}. 


\section{Learning Invariant Representations and Risks}
\label{sec:LearningRepresentations}

Motivated by the generalization error bounds in Theorem~\ref{eq:classification_upper_bound} and Theorem~\ref{eq:regression_upper_bound} in Sec.~\ref{sec:bound}, in this section we propose our bound minimization algorithm LIRR. Since the last two terms reflect the noise level, complexity measures and error caused by finite samples, respectively, we then hope to optimize the upper bound by minimizing the first four terms. The first two terms are the convex combination of empirical errors of $h$ on $S$ and $T$, which can be optimized with the labeled source and target data. The third term measures the distance of representations between the source and target domains, which is a good inspiration for us to learn the \emph{invariant representation} across domains. The fourth term corresponds to the distance of the optimal classifiers between $S$ and $T$. To minimize this term, the model is forced to learn the data representations that induce the same optimal predictors for both source and target domains, which exactly corresponds to the principle of \emph{invariant risk minimization}~\cite{arjovsky2019invariant}. Several efforts in the fairness representation area~\cite{song2019learning, steinberg2020fast} and domain generalization~\cite{krueger2020out, zhao2020domain} area have proposed similar ideas of invariant risk. However, LIRR is the first work to combine both \emph{invariant representation} and \emph{invariant risk minimization} for applications in semisupervised domain adaptation.



\subsection{Information Theoretic Interpretation}
To better understand why the bound minimization strategy can solve the intrinsic problems of Semi-DA, in what follows we provide interpretations from an information-theoretic perspective.
\paragraph{Invariant Representations}
Learning invariant representations corresponds to minimizing the third term of the bound Theorem~\ref{eq:classification_upper_bound} and bound Theorem~\ref{eq:regression_upper_bound}. We consider a feature transformation $Z = g(X)$ that can obtain the invariant representation $Z$ from input $X$. The invariance on representations can be described as achieving statistical independence $D \perp Z$, where $D$ stands for the domain index. This independence is equivalent to the minimization of mutual information $I(D; Z)$. To see this, if $I(D; Z) = 0$, then $\dist_S(Z) = \dist_T(Z)$, so the third term in the bounds will vanish. Intuitively, this means that by looking at the representations $Z$, even a well-trained domain classifier $\mathcal{C}(\cdot)$ cannot correctly guess the domain index $D$. 
\paragraph{Invariant Risks}
Learning invariant risks corresponds to minimizing the fourth term of the bound Theorem~\ref{eq:classification_upper_bound} and bound Theorem~\ref{eq:regression_upper_bound}. Inspired by ~\cite{arjovsky2019invariant}, we want to identify a subset of feature representations through feature transformation $Z = g(X)$ that best supports an invariant optimal predictor for source and target domains. That means the identified feature representation $Z=g(X)$ can induce the same optimal predictors. This objective can be interpreted with a conditional independence $D \perp Y \mid Z$, which is equivalent to minimizing $I(D; Y \mid Z)$. To see this, when the conditional mutual information of $I(D; Y \mid Z)$ equals $0$, the two conditional distributions $\Pr_S(Y\mid Z)$ and $\Pr_T(Y\mid Z)$ coincide with each other. As a result, the Bayes optimal predictors, which only depend on the conditional distributions of $Y\mid Z$, become the same across domains, so the fourth term in our bounds Theorem~\ref{eq:classification_upper_bound}, Theorem~\ref{eq:regression_upper_bound} will vanish. 

In summary, our learning objective on invariant representations and invariant risks are achievable with the joint minimization of $I(D; Z)$ and $I(D; Y \mid Z)$. It is instructive to present the integrated form as in Eq.~\ref{eq:integrated_mutual_information}. In words, the integrated form suggests the independence of $D \perp (Y, Z)$. We regard the independence as an intrinsic objective for domain adaptation since it implies an alignment of the joint distributions over $(Y, Z)$ across domains, as opposed to only the marginal distributions over $Z$ in existing works.
\begin{equation}
    I(D; Y, Z) = \underbrace{I(D; Z)}_{\text{Invariant Representation}} + \underbrace{I(D; Y \mid Z)}_{\text{Invariant Risk}}.
    \label{eq:integrated_mutual_information}
\end{equation}


\subsection{Algorithm Design}
To learn invariant representations, that is achieving marginal independence of $Y \perp Z$ and minimization on $\min I(Y; Z)$, we adopt the adversarial training method as in~\cite{ganin2016domain}. The invariant representation objective focuses on learning the feature transformation $g(\cdot)$ to obtain the \emph{invariant representations} from input $X$, which can fool the domain classifier $\mathcal{C}$. This part of the objective function can be described as in Eq.~\ref{eq:dann_loss}.
\begin{equation}
\begin{aligned}
    \mathcal{L}_{\text{rep}}(g, \mathcal{C}) = \ &\mathbb{E}_{X \sim \mathcal{D}_S(X)}[\log(\mathcal{C}(g(X)))] \\
    &+ \ \mathbb{E}_{X \sim \mathcal{D}_T(X)} [\log (1 - \mathcal{C}(g(X)))].
    \label{eq:dann_loss}
\end{aligned}
\end{equation}

To learn invariant risks, that is achieving conditional independence of $D \perp Y \mid Z$, we resort to the conditional mutual information minimization on $I(D; Y \mid Z)$, and further convert $\min I(D; Y \mid Z)$ objective to the minimization of the difference between the following conditional entropies:
\begin{equation}
    I(D; Y \mid Z) = H(Y \mid Z) - H(Y \mid D, Z).
    \label{eq:decomposation}
\end{equation}
The following proposition gives a variational form of the conditional entropy as infimum over a family of cross-entropies, where $L$ denotes the cross-entropy loss.
\begin{proposition}[\cite{farnia2016minimax}]
$H(Y \mid Z) = \inf_f\mathbb{E}[L(Y; f(Z))]$.
\end{proposition}
Using the above variational form, the minimization of the conditional entropies could be transformed to a minimization of the cross-entropy losses of domain-invariant predictor $f_i$ and domain-dependent predictor $f_d$. The learning objective of the two predictors can be shown as in Eq.~\ref{eq:ent_condition_on_z} and Eq.~\ref{eq:ent_condition_on_z_d}, respectively. Notice that the domain-dependent loss $\mathcal{L}_{d}$ should be no greater than the domain-invariant loss $\mathcal{L}_{i}$, because of the additional domain information. 

\begin{equation}
\begin{aligned}
    \min\limits_{g, f_i}\mathcal{L}_{i} = \mathbb{E}_{(x, y) \sim \mathcal{D}_S, \mathcal{D}_{\tilde{T}}}[L(y, f_{i}(g(x)))],
\label{eq:ent_condition_on_z}
\end{aligned}
\end{equation}

\begin{equation}
\begin{aligned}
    \min\limits_{g, f_d}\mathcal{L}_{d} = \mathbb{E}_{d \sim D} \mathbb{E}_{(x, y) \sim \mathcal{D}_{d}}[L(y, f_{d}(g(x), d))].
\label{eq:ent_condition_on_z_d}
\end{aligned}
\end{equation}

Hence, the overall learning objective of Eq.~\ref{eq:decomposation} can be re-written with the following loss functions.

\begin{equation}
\begin{aligned}
    \min\limits_{g, f_i} & \max\limits_{f_d} \; \mathcal{L}_{\text{risk}} = \mathcal{L}_{i} + \lambda_{risk}( \mathcal{L}_{i} - \mathcal{L}_{d}).
\label{eq:lirr_learning_objective}
\end{aligned}
\end{equation}

The first term of Eq.~\ref{eq:lirr_learning_objective} regards to the supervised training on source and target labeled data; the second term regards to approaching the minimization objective of $H(Y \mid Z) - H(Y \mid D, Z)$, as well as achieving the predictions' invariance between $f_i$ and $f_d$ over the same representation $z$. If we take the example of binary classification of bear and dog as in Eq .~\ref{fig:problem}, if $f_i$ and $f_d$ have their prediction of bear according to a proper representation of animal's shape, then any domain information will not contribute to the prediction, thus the predictor captures the invariant part and achieves \emph{invariant risks}.

In general, as the factorization in Eq.~\ref{eq:integrated_mutual_information} suggests, in order to achieve improved adaptation performance by minimizing the accuracy discrepancy between domains, we need to enforce the joint independence of $\left(Y, Z\right) \perp D$ by learning feature transformation $g$. To achieve it, we propose our learning objective of LIRR as in Eq.~\ref{eq:minmax_objective}, where $\lambda_{\text{risk}}$ and $\lambda_{\text{rep}}$ are set to 1 by default.
\begin{equation}
\begin{aligned}
    \min\limits_{g, f_i} \max\limits_{\mathcal{C}, f_d} \; & \mathcal{L}_{\text{LIRR}}(g, f_i, f_d, \mathcal{C}) \\ 
    &= \mathcal{L}_{\text{risk}}(g, f_i, f_d) + \ \lambda_{\text{rep}} \mathcal{L}_{\text{rep}}(g, \mathcal{C}).
    \label{eq:minmax_objective}
\end{aligned}
\end{equation}
At a high level, the first term $\mathcal{L}_{\text{risk}}(g, f_i, f_d)$ in the above optimization formulation stems from the minimization of $I(Y; D\mid Z)$, and the second term $\mathcal{L}_{\text{rep}}(g, \mathcal{C})$ is designed to minimize $I(D; Z)$.

\section{Experiments}
To empirically corroborate the effectiveness of LIRR, in this section we conduct experiments on both classification and regression tasks under the setting of Semi-DA and compare LIRR to existing methods. We first introduce the experimental settings, and then present analysis to the experimental results. We also provide ablation study for the experiments on both classification and regression tasks. More experimental settings, implementation details, and results are discussed in the Appendix.

\subsection{Image Classification}
\label{sec:experiment_img_cls}
\paragraph{Datasets} To verify the effectiveness of LIRR on image classification problems, we conduct experiments on NICO~\cite{he2020towards}, VisDA2017~\cite{peng2017visda}, OfficeHome~\cite{venkateswara2017deep}, and DomainNet~\cite{peng2019moment} datasets. \textit{NICO} is dedicatedly designed for \textbf{O.O.D.} (out-of-distribution) image classification. It has two superclasses \emph{animal} and \emph{vehicle}, and each superclass contains different environments\footnote{For \emph{animal}, we sample 8 classes from environments \emph{grass} and \emph{snow} as two domains. For \emph{vehicle}, we sample 7 classes from environments \emph{sunset} and \emph{beach} as two domains.}, \emph{e.g.} bear on grass or snow.  \textit{VisDA2017} contains Train (T) domain and Validation (V) domain with 12 classes in each domain. \textit{Office-Home} includes four domains: RealWorld (RW), Clipart (C), Art (A), and Product (P), with 65 classes in each domain. \textit{DomainNet} is the largest domain adaptation dataset for image classification with over 600k images from 6 domains: Clipart (C), Infograph (I), Painting (P), Quickdraw (Q), Real (R), and Sketch (S), with 345 classes in each domain. For each dataset, we randomly pick source-target pairs for evaluation. To meet the setting of Semi-DA, we randomly select a small ratio ($1\%$ or $5\%$) of the target data as labeled target samples for training. More information about datasets will be detailed in appendix.

\paragraph{Baselines} We compare our approach with the following representative domain adaptation methods:
\textit{DANN}~\cite{ganin2016domain}, \textit{CDAN}~\cite{long2018conditional}, \textit{IRM}~\cite{arjovsky2019invariant}, \textit{ADR}~\cite{saito2017adversarial}, and \textit{MME}~\cite{saito2019semi}; \textit{S+T}, a model trained with the labeled source and the few labeled target samples without using unlabeled target samples; and \textit{Full T}, a model trained with the fully labeled target.
All these methods are implemented and evaluated under the Semi-DA setting. 

\begin{table*}[tb]
\renewcommand\tabcolsep{4 pt} 
\centering\footnotesize
\caption{Accuracy (\%) comparison (higher means better) on \textbf{NICO}, \textbf{OfficeHome}, \textbf{DomainNet}, and \textbf{VisDA2017} with 1\% (above) and 5\% (below) labeled target data (mean $\pm$ std). Highest accuracies are highlighted in bold.}
\resizebox{\linewidth}{!}{%
\begin{tabular}{l|cc|cc|ccc|ccc|c}
\toprule
1\% \scriptsize{labeled target} & \multicolumn{2}{c|}{\centering NICO Animal}& \multicolumn{2}{c|}{\centering NICO Traffic} & \multicolumn{3}{c|}{\centering OfficeHome}& \multicolumn{3}{c|}{\centering Domainnet} & \multicolumn{1}{c}{\centering VisDA2017} \\
\hline
 Method & \scriptsize{Grass to Snow} & \scriptsize{Snow to Grass} & \scriptsize{Sunset to Beach} & \scriptsize{Beach to Sunset} &\scriptsize{Art to Real} & \scriptsize{Real to Prod.} & \scriptsize{Prod. to Clip.} & \scriptsize{Real to Clip.} & \scriptsize{Sketch to Real} & \scriptsize{Clip. to Sketch}  & \scriptsize{Train to Val.} \\
\hline
S+T & 70.06\scriptsize{$\pm$2.14} & 80.08\scriptsize{$\pm$1.21} & 71.37\scriptsize{$\pm$1.54} & 70.07\scriptsize{$\pm$1.28} & 69.20\scriptsize{$\pm$0.15 }& 74.63\scriptsize{$\pm$0.13 }& 48.65\scriptsize{$\pm$0.12 } & 48.37\scriptsize{$\pm$0.08 } & 57.44\scriptsize{$\pm$0.07 }& 44.16\scriptsize{$\pm$0.05 } & 76.17\scriptsize{$\pm$0.15 } \\
\hline
DANN & 83.80\scriptsize{$\pm$1.73} & 81.57\scriptsize{$\pm$1.51} & 72.69\scriptsize{$\pm$1.35} & 72.03\scriptsize{$\pm$1.05} & 72.20\scriptsize{$\pm$0.23 }& 78.13\scriptsize{$\pm$0.26 }& 52.47\scriptsize{$\pm$0.21 } & 51.53\scriptsize{$\pm$0.19 } & 60.23\scriptsize{$\pm$0.15 }& 46.36\scriptsize{$\pm$0.15 }&  78.91\scriptsize{$\pm$0.25 } \\
CDAN & 82.33\scriptsize{$\pm$0.59} & 78.25\scriptsize{$\pm$0.74} & 75.53\scriptsize{$\pm$0.55} & 74.31\scriptsize{$\pm$0.47} & 72.98\scriptsize{$\pm$0.33 }& 79.15\scriptsize{$\pm$0.31 }& 53.80\scriptsize{$\pm$0.33 } & 50.67\scriptsize{$\pm$0.25 }& 60.53\scriptsize{$\pm$0.23 }& 44.66\scriptsize{$\pm$0.22 } & 80.23\scriptsize{$\pm$0.41 } \\
ADR & 73.06\scriptsize{$\pm$1.20} & 76.74\scriptsize{$\pm$0.89} & 72.85\scriptsize{$\pm$0.95} & 69.47\scriptsize{$\pm$0.81} & 70.55\scriptsize{$\pm$0.27 }& 76.62\scriptsize{$\pm$0.28 }& 49.47\scriptsize{$\pm$0.31 } & 49.94\scriptsize{$\pm$0.21 }& 59.63\scriptsize{$\pm$0.22 }& 44.73\scriptsize{$\pm$0.21 } & 80.40\scriptsize{$\pm$0.36 } \\
IRM & 78.55\scriptsize{$\pm$0.34} & 78.27\scriptsize{$\pm$0.51} & 64.58\scriptsize{$\pm$2.41} & 69.10\scriptsize{$\pm$2.36} & 71.13\scriptsize{$\pm$0.25 }& 77.60\scriptsize{$\pm$0.24 }& 51.53\scriptsize{$\pm$0.21 } & 51.86\scriptsize{$\pm$0.13 }& 58.04\scriptsize{$\pm$0.12 }& 46.96\scriptsize{$\pm$0.15 } & 80.79\scriptsize{$\pm$0.27 } \\

MME & 87.12\scriptsize{$\pm$0.76} & 79.52\scriptsize{$\pm$0.43} & 78.69\scriptsize{$\pm$0.86} & 74.21\scriptsize{$\pm$0.78} & 72.66\scriptsize{$\pm$0.18 }& 78.07\scriptsize{$\pm$0.17 }& 52.78\scriptsize{$\pm$0.16 } & 51.04\scriptsize{$\pm$0.12 }& 60.35\scriptsize{$\pm$0.12 }& 45.09\scriptsize{$\pm$0.14 } & 80.52\scriptsize{$\pm$0.35 } \\

LIRR & 86.80\scriptsize{$\pm$0.61} & 84.78\scriptsize{$\pm$0.53} & 71.85\scriptsize{$\pm$0.58} & 72.04\scriptsize{$\pm$0.75} & 73.12\scriptsize{$\pm$0.19 }& 79.58\scriptsize{$\pm$0.22 }& \textbf{54.33}\scriptsize{$\pm$0.24 } & 52.39\scriptsize{$\pm$0.15 }& 61.20\scriptsize{$\pm$0.10 }& 47.31\scriptsize{$\pm$0.11 } & 81.67\scriptsize{$\pm$0.22 } \\
LIRR+CosC & \textbf{89.67}\scriptsize{$\pm$0.72} & \textbf{89.73}\scriptsize{$\pm$0.68} & \textbf{81.00}\scriptsize{$\pm$0.89} & \textbf{79.98}\scriptsize{$\pm$0.95} & \textbf{73.62}\scriptsize{$\pm$0.21 }& \textbf{80.20}\scriptsize{$\pm$0.23 }& 53.84\scriptsize{$\pm$0.19 } & \textbf{53.42}\scriptsize{$\pm$0.09 }& \textbf{61.79}\scriptsize{$\pm$0.11 }& \textbf{47.83}\scriptsize{$\pm$0.10 } & \textbf{82.31}\scriptsize{$\pm$0.21 } \\
\hline
Full T & 94.52\scriptsize{$\pm$0.74} & 97.98\scriptsize{$\pm$0.23} & 99.80\scriptsize{$\pm$0.87} & 97.64\scriptsize{$\pm$0.96} & 83.67\scriptsize{$\pm$0.12 }& 91.42\scriptsize{$\pm$0.05 }& 78.27\scriptsize{$\pm$0.23 } & 72.40\scriptsize{$\pm$0.05 }& 77.11\scriptsize{$\pm$0.07 }& 62.66\scriptsize{$\pm$0.07 } & 89.56\scriptsize{$\pm$0.14 } \\
\hline
\hline
\toprule
5\% \scriptsize{labeled target} & \multicolumn{2}{c|}{\centering NICO Animal}& \multicolumn{2}{c|}{\centering NICO Traffic} & \multicolumn{3}{c|}{\centering OfficeHome}& \multicolumn{3}{c|}{\centering Domainnet} & \multicolumn{1}{c}{\centering VisDA2017} \\
\hline
 Method & \scriptsize{Grass to Snow} & \scriptsize{Snow to Grass} & \scriptsize{Sunset to Beach} & \scriptsize{Beach to Sunset} &\scriptsize{Art to Real} & \scriptsize{Real to Prod.} & \scriptsize{Prod. to Clip.} & \scriptsize{Real to Clip.} & \scriptsize{Sketch to Real} & \scriptsize{Clip. to Sketch}  & \scriptsize{Train to Val.} \\
\hline
S+T & 75.83\scriptsize{$\pm$1.89} & 83.38\scriptsize{$\pm$1.23} & 86.45\scriptsize{$\pm$1.08} & 86.13\scriptsize{$\pm$0.87} & 72.10\scriptsize{$\pm$0.13} & 78.84\scriptsize{$\pm$0.12} & 54.51\scriptsize{$\pm$0.10} & 59.80\scriptsize{$\pm$0.13} & 66.14\scriptsize{$\pm$0.11} & 51.71\scriptsize{$\pm$0.09} & 82.87\scriptsize{$\pm$0.12} \\
\hline
DANN & 76.13\scriptsize{$\pm$0.73} & 84.61\scriptsize{$\pm$1.21} & 84.13\scriptsize{$\pm$1.20} & 87.50\scriptsize{$\pm$1.09} & 75.47\scriptsize{$\pm$0.22} & 80.41\scriptsize{$\pm$0.21} & 59.37\scriptsize{$\pm$0.20} & 61.31\scriptsize{$\pm$0.14} & 68.21\scriptsize{$\pm$0.20} & 52.78\scriptsize{$\pm$0.22} & 83.95\scriptsize{$\pm$0.10} \\
CDAN & 82.33\scriptsize{$\pm$0.59} & 83.08\scriptsize{$\pm$2.13} & 86.97\scriptsize{$\pm$0.47} & 87.50\scriptsize{$\pm$0.56} & 74.92\scriptsize{$\pm$0.29} & 80.57\scriptsize{$\pm$0.33} & 59.14\scriptsize{$\pm$0.31} & 62.18\scriptsize{$\pm$0.22} & 68.49\scriptsize{$\pm$0.19} & 53.77\scriptsize{$\pm$0.21} & 83.31\scriptsize{$\pm$0.32} \\
ADR & 80.36\scriptsize{$\pm$0.31} & 80.97\scriptsize{$\pm$0.98} & 84.50\scriptsize{$\pm$0.91} & 75.29\scriptsize{$\pm$0.87} & 75.47\scriptsize{$\pm$0.27} & 79.27\scriptsize{$\pm$0.26} & 58.24\scriptsize{$\pm$0.27} & 61.22\scriptsize{$\pm$0.38} & 67.96\scriptsize{$\pm$0.37} & 53.19\scriptsize{$\pm$0.32} & 83.57\scriptsize{$\pm$0.43} \\
IRM & 81.57\scriptsize{$\pm$1.01} & 84.29\scriptsize{$\pm$1.10} & 85.71\scriptsize{$\pm$2.20} & 83.61\scriptsize{$\pm$2.17} & 74.71\scriptsize{$\pm$0.21} & 79.67\scriptsize{$\pm$0.25} & 58.98\scriptsize{$\pm$0.22} & 60.69\scriptsize{$\pm$0.30} & 67.81\scriptsize{$\pm$0.28} & 52.31\scriptsize{$\pm$0.25} & 82.62\scriptsize{$\pm$0.29} \\
MME & 87.80\scriptsize{$\pm$0.87} & 85.50\scriptsize{$\pm$0.95} & 92.02\scriptsize{$\pm$0.85} & 90.76\scriptsize{$\pm$0.81} & 75.24\scriptsize{$\pm$0.22} & 82.45\scriptsize{$\pm$0.18} & 61.75\scriptsize{$\pm$0.19} & 62.31\scriptsize{$\pm$0.11} & 69.02\scriptsize{$\pm$0.18} & 53.88\scriptsize{$\pm$0.14} & 84.12\scriptsize{$\pm$0.22} \\
LIRR & 85.90\scriptsize{$\pm$0.98} & 85.24\scriptsize{$\pm$0.73} & 90.77\scriptsize{$\pm$0.42} & 88.90\scriptsize{$\pm$0.39} & 76.14\scriptsize{$\pm$0.18} & \textbf{83.64}\scriptsize{$\pm$0.21} & 62.61\scriptsize{$\pm$0.17} & 62.74\scriptsize{$\pm$0.21} & 69.35\scriptsize{$\pm$0.13} & 54.05\scriptsize{$\pm$0.17} & 84.47\scriptsize{$\pm$0.19} \\
LIRR+CosC & \textbf{88.97}\scriptsize{$\pm$0.45} & \textbf{88.22}\scriptsize{$\pm$0.55} & \textbf{92.70}\scriptsize{$\pm$0.87} & \textbf{91.50}\scriptsize{$\pm$1.05} & \textbf{76.63}\scriptsize{$\pm$0.19} & 83.45\scriptsize{$\pm$0.22} & \textbf{62.84}\scriptsize{$\pm$0.23} & \textbf{63.03}\scriptsize{$\pm$0.17} & \textbf{69.52}\scriptsize{$\pm$0.09} & \textbf{54.44}\scriptsize{$\pm$0.12} & \textbf{85.06}\scriptsize{$\pm$0.17} \\
\hline
Full T & 94.52\scriptsize{$\pm$0.74} & 97.98\scriptsize{$\pm$0.23} & 99.80\scriptsize{$\pm$0.87} & 97.64\scriptsize{$\pm$0.96} & 83.67\scriptsize{$\pm$0.12} & 91.42\scriptsize{$\pm$0.05} & 78.27\scriptsize{$\pm$0.23} & 72.40\scriptsize{$\pm$0.05} & 77.11\scriptsize{$\pm$0.07} & 62.66\scriptsize{$\pm$0.07} & 89.56\scriptsize{$\pm$0.14} \\
\hline
\end{tabular}
}
\label{tab:classification}
\end{table*}

\begin{table*}[tb]
\centering
\caption{Mean absolute error (MAE, lower means better) comparison on \textbf{WebCamT} with 1\% and 5\% labeled target data (mean $\pm$ std). The best is emphasized in bold.}
\label{tab:city_cam}

\begin{tabular}{l|cc|cc|cc}
\toprule
\multirow{2}{*}{Method} & \multicolumn{2}{c|}{253 to 398} & \multicolumn{2}{c|}{170 to 398} & \multicolumn{2}{c}{511 to 398} \\ \cline{2-7} 
& 1\% & 5\% & 1\% & 5\% & 1\% & 5\%  \\
\hline
S+T & 3.20\scriptsize{$\pm$0.03 }& 2.42\scriptsize{$\pm$0.02 }& 3.12\scriptsize{$\pm$0.02 }& 2.07\scriptsize{$\pm$0.01 }& 3.45\scriptsize{$\pm$0.02 }& 2.82\scriptsize{$\pm$0.04} \\
\hline
ADDA & 3.13\scriptsize{$\pm$0.01} & 2.34\scriptsize{$\pm$0.03} & 3.05\scriptsize{$\pm$0.03} & 2.05\scriptsize{$\pm$0.01} & 2.87\scriptsize{$\pm$0.03} &  2.45\scriptsize{$\pm$0.02} \\
DANN & 3.08\scriptsize{$\pm$0.02 }& 2.38\scriptsize{$\pm$0.02 }& 3.01\scriptsize{$\pm$0.04 }& 2.01\scriptsize{$\pm$0.02 }& 2.95\scriptsize{$\pm$0.03 }& 2.41\scriptsize{$\pm$0.04} \\
IRM & 3.11\scriptsize{$\pm$0.02 }& 2.27\scriptsize{$\pm$0.03 }& 2.91\scriptsize{$\pm$0.02 }& 2.02\scriptsize{$\pm$0.01 }& 2.89\scriptsize{$\pm$0.05 }& 2.33\scriptsize{$\pm$0.03} \\
LIRR &  \textbf{2.96}\scriptsize{$\pm$0.02 }&  \textbf{2.13}\scriptsize{$\pm$0.01 }&  \textbf{2.84}\scriptsize{$\pm$0.01 }&  \textbf{1.98}\scriptsize{$\pm$0.02 }&  \textbf{2.80}\scriptsize{$\pm$0.03 }&  \textbf{2.25}\scriptsize{$\pm$0.01} \\
\hline
Full T & 1.68\scriptsize{$\pm$0.01 }& 1.68\scriptsize{$\pm$0.01 }& 1.68\scriptsize{$\pm$0.01 }& 1.68\scriptsize{$\pm$0.01 }& 1.68\scriptsize{$\pm$0.01 }& 1.68\scriptsize{$\pm$0.01} \\
\hline
\end{tabular}
\label{tab:regression}
\end{table*}

\subsection{Traffic Counting Regression}

\paragraph{Datasets} To verify the effectiveness of LIRR on regression problems, we conduct experiments on WebCamT dataset~\cite{zhang2017understanding} for the Traffic Counting Regression task. WebCamT has 60,000 traffic video frames annotated with vehicle bounding boxes and counts, collected from 16 surveillance cameras with different locations and recording time. We pick three source-target pairs with different visual similarities: 253$\rightarrow$398, 170$\rightarrow$398, 511$\rightarrow$398 (digit denotes camera ID). 

\paragraph{Baselines} The baseline models for this task are generally aligned with our classification experiments except the methods that can not be applied to the regression task (e.g. \textit{MME}, \textit{ADR}, and \textit{CDAN}). Thus, for the traffic counting regression task, we compare with the baseline methods: \textit{ADDA}~\cite{tzeng2017adversarial}, \textit{DANN}, \textit{IRM}, \textit{S+T}, and \textit{FullT}.

\begin{figure*}[tb]
    \centering
    \includegraphics[width=1\linewidth]{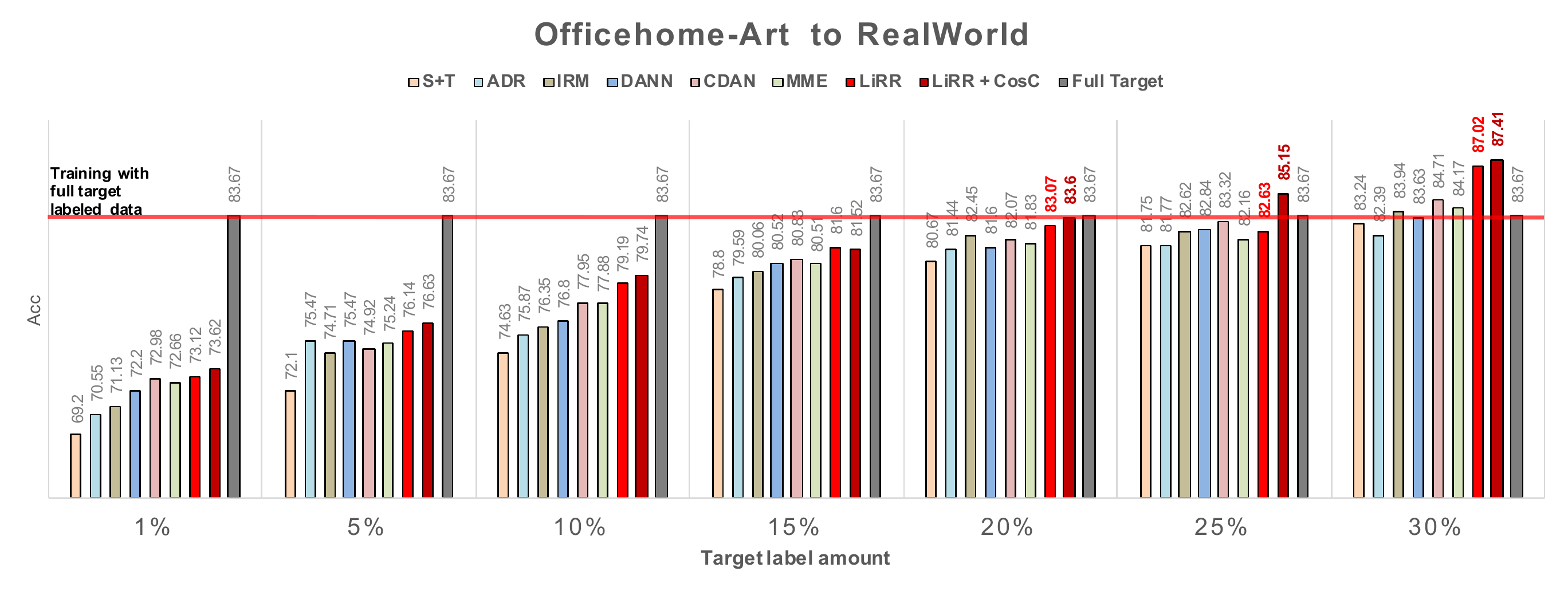}
    \vspace{-0.5cm}
    \caption{Performance comparison with increasing number of labeled target data, from Domain Art to RealWorld on Officehome dataset. X axis: the ratio of labeled target data; Y axis: accuracy.}
    \label{fig:increasing_with_target_amount}
    \vspace{-0.5cm}
\end{figure*}

\subsection{Experimental Results Analysis}
\label{sec:experiment_results}
\paragraph{Classification Tasks}
The classification results are shown in Table~\ref{tab:classification} with 1\% and 5\% labeled target data. LIRR outperforms the baselines on all the five adaptation datasets, which consistently indicates its effectiveness. As our learning objective suggests, LIRR can be viewed as achieving $D \perp (Y, Z)$, which combines the benefits of achieving $D \perp Y$ and $D \perp Y \mid Z$. In contrast, DANN, CDAN, and ADDA can be viewed as only achieving $D \perp Z$ or its variant form; and IRM can be viewed as an approximation to achieve $D \perp Y \mid Z$ using gradient penalty. LIRR outperforms all these methods on different datasets with 1\% or 5\% labeled target data, demonstrating simultaneously learning invariant representations and risks achieves better generalization for domain adaptation than only learning one of them. Such results are consistent with our theoretical analysis and algorithm design objective.
Besides, when applying LIRR along with the cosine classifier (\textit{CosC}) module, which is also used in MME, the performance further outperforms MME by a larger margin.

\vspace{-1cm}
\paragraph{Regression Tasks}
The traffic counting regression results are shown in Table~\ref{tab:regression} with 1\% and 5\% labeled target data. 
The superiority of LIRR over baseline methods is supported by its lowest MAE on all the settings. 
DANN and ADDA are the representative methods of learning invariant representations, while IRM is the representative method of learning invariant risks. Both DANN, ADDA, and IRM achieve lower error than S+T, which means learning invariant representations or invariant risks can benefit Semi-DA to some extent on the regression task. 
Similar with the observations from the classification experiments, LIRR outperforms both DANN, ADDA, and IRM, demonstrating simultaneously learning invariant representations and risks achieves better adaptation than only aligning one of them. 

\subsection{Ablation Study}
\paragraph{Comparisons with Optimizing Single Invariant Objective} 
As pointed out in Sec.~\ref{sec:experiment_results}, LIRR is simultaneously learning invariant representations and risks, while DANN, CDAN, ADDA can be viewed as only achieving invariant representations or its variant forms, and IRM is an approximation to solely achieve invariant risks. From the results on both classification and regression tasks, we can further acknowledge the importance of simultaneously optimizing these two invariant items together. As shown in Table~\ref{tab:classification} and ~\ref{tab:regression}, all the methods that only minimize one single invariant objective perform worse than LIRR, indicating our method is effective and consistent to the theoretical results.
\vspace{-0.6cm}
\paragraph{Increasing Proportions of Labeled Target Data} 
Revisiting Theorem~\ref{eq:classification_upper_bound} and Theorem~\ref{eq:regression_upper_bound}, we know that as the proportion of the labeled target data rises, the upper bound of $\epsilon_{T}(h)$ gets tighter. Accordingly, the margin between LIRR and other methods becomes larger, as shown in Fig.~\ref{fig:increasing_with_target_amount}. 
Another riveting observation from Fig.~\ref{fig:increasing_with_target_amount} is, LIRR and its variant LIRR+CosC achieve better performance than the oracle by large margin with 25\% or 30\% labeled target data. Stunning but plausible, with source and a few labeled target data, LIRR can learn more robust representations and achieve better performance on the target, comparing with the model trained by the fully labeled target data.

\paragraph{Cosine Classifier} 
As introduced in \cite{saito2019semi}, cosine classifier is proved to be helpful for improving the model's performance on Semi-DA. As shown in Table.~\ref{tab:classification}, the same phenomenon can be found when comparing the performance of LIRR and LIRR+CosC. For almost all the cases, LIRR plus cosine classifier module achieves higher accuracy than LIRR alone. 

\subsection{Visualization Results}
\begin{figure}[t]
    \centering
    \includegraphics[width=1\linewidth]{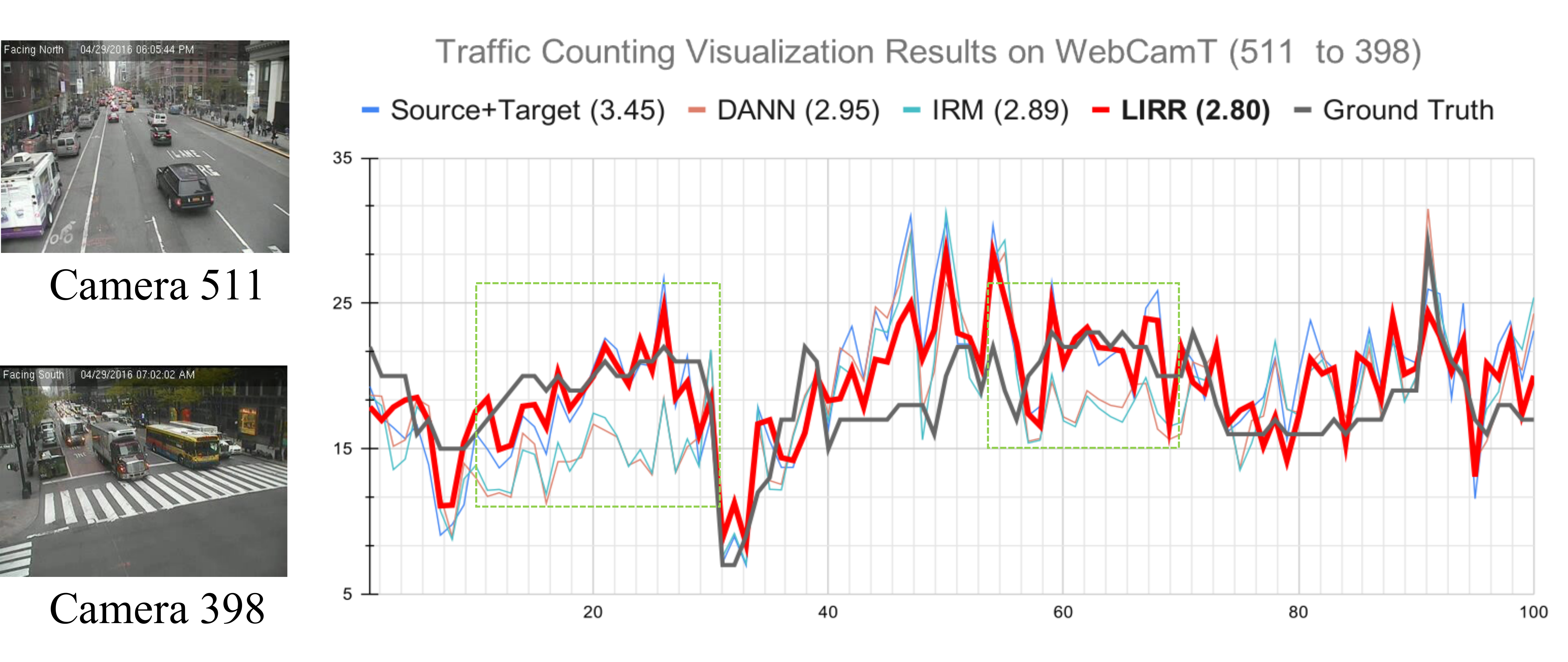}
    \caption{The line chart of the regression results of different DA methods on Camera 511 to 398, WebCamT.}
    \label{fig:counting_vis}
\end{figure}

\begin{figure}[t]
    \centering
    \includegraphics[width=1\linewidth]{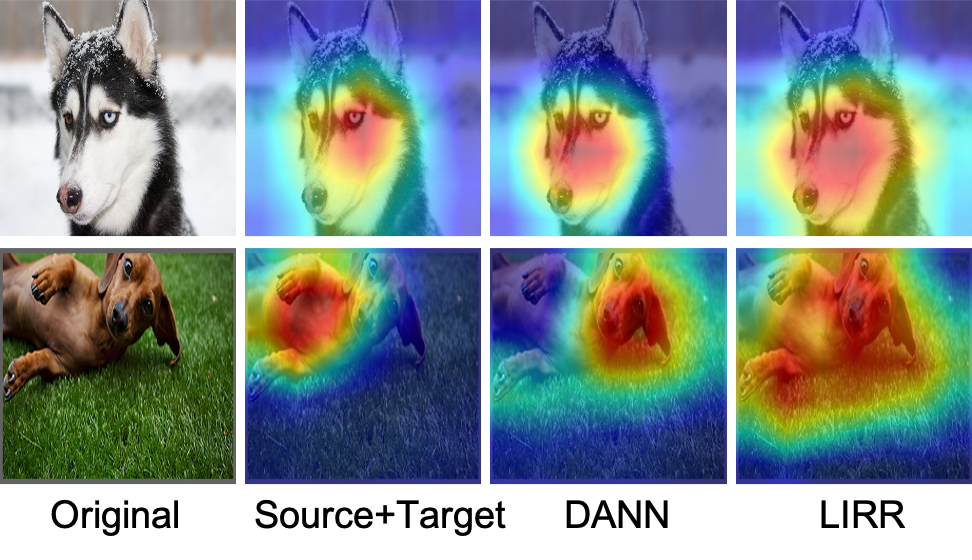}
    \caption{Grad-CAM~\cite{selvaraju2017grad} results of different model. LIRR appropriately captures the invariant part of the same object in different domains, \emph{e.g.} the shape of horse leads to invariant prediction across snow and grass domain.}
    \label{fig:cvpr_gradcam}
    \vspace{-0.5cm}
\end{figure}

Fig.~\ref{fig:counting_vis} visualizes the counting results of different algorithms on Camera 511 to 398 scenario, WebCamT. The red line represents the LIRR method we proposed while the black line represents the gt count. It's rather clear to see that LIRR have a better ability of cross domain regression fitting than other methods, especially the area within the green bounding box with dot lines. In order to vividly showcase the learned feature representation which supports the invariant risks across domains. We employ Grad-CAM~\cite{selvaraju2017grad} to visualize the most influential part in prediction in Fig~\ref{fig:cvpr_gradcam}.




\section{Conclusion}
In this paper, we argue that, compared with UDA, the setting of Semi-DA is more realistic and enjoys broader practical applications with potentially better utility. To this end, in this paper we propose the first finite-sample generalization bounds for both classification and regression problems under Semi-DA. Our results shed new light on Semi-DA by suggesting a principled way of simultaneously learning invariant representations and risks across domains, leading to a bound minimization algorithm - LIRR. Extensive experiments on real-world datasets, including both image classification and traffic counting tasks, demonstrate the effectiveness.

\small
\bibliographystyle{ieeetr}
\bibliography{egbib}

\def\cvprPaperID{11397} 
\def\confYear{CVPR 2021}
\setcounter{section}{0}
\section{Proof}
In this section, we provide a detailed proof of Thm. 4.1 and Thm. 4.2 in sequence. 
\subsection{Proof of classification bound}
Before we reach the proof to the main theorem, we first prove the following lemmas for each theorem. With the notations introduced in Sec. 4, we introduce the following lemmas that will be used in proving the main theorem:

\begin{restatable}{lemma}{first_lemma_cls}
\label{lemma:lemma_a1}
[\cite{blitzer2008learning}] Let $h \in \mathcal{H}:=\{h: \hdata \rightarrow\{0, 1\}\}$, where $VCdim(\HH) = d$, and for any distribution $\mathcal{D}_S(Z)$, $\mathcal{D}_T(Z)$ over $\hdata$, then
\begin{align*}
    |\epsilon_S(h) - \epsilon_T(h)| \leq  d_{\HH \Delta \HH} (\mathcal{D}_S(Z), \mathcal{D}_T(Z))
\end{align*}
\end{restatable}

\begin{lemma}
\label{lemma:lemma_a2}
Let $h \in \mathcal{H}:=\{h: \hdata \rightarrow\{0, 1\}\}$, where $VCdim(\HH) = d$, and for any distribution $\mathcal{D}_S(Z)$, $\mathcal{D}_T(Z)$ over $\hdata$. Let the noises on the source and target are defined as $n_S \defeq \Exp_S[|Y - f_S(Z)|]$ and $n_T \defeq \Exp_T[|Y - f_T(Z)|]$, where $f : \hdata \rightarrow [0, 1]$ is the conditional mean function, i.e., $f(Z) = \Exp[Y | Z]$ then we have:
\begin{align*}
& \big|\eps_S(h)-\eps_T(h)\big| \leq |n_S + n_T| +  d_{\HH \Delta \HH} (\mathcal{D}_S(Z), \mathcal{D}_T(Z))  \\
&+ \min\{\Exp_S[|f_S(Z) - f_T(Z)|], \Exp_T[|f_S(Z) - f_T(Z)|]\}
\end{align*}
\end{lemma}
\begin{proof}
To begin with, we first show that for the source domain, $\eps_S(h)$ cannot be too large if $h$ is close to the optimal classifier $f_S$ on source domain for $\forall h\in\HH$:
\begin{align*}
& |\eps_S(h) - \Exp_S[|h(Z) - f_S(Z)|]| \\
& =\big|\Exp_S[|h(Z) - Y|] - \Exp_S[|h(Z) - f_S(Z)|]\big|  \\
                      &\leq \Exp_S\left[\big||h(Z) - Y| - |f_S(Z) - h(Z)|\big|\right] \\
                      &\leq \Exp_S[|Y - f_S(Z)|] \\
                      &= n_S
\end{align*}
Similarly, we also have an analogous inequality hold on the target domain:
\begin{equation*}
    |\eps_T(h) - \Exp_T[|h(Z) - f_T(Z)|]| \leq n_T.
\end{equation*}
Combining both inequalities above, yields:
\begin{align*}
    \eps_S(h) \in & \; [\Exp_S[|h(Z) - f_S(Z)|] - n_S, \\
     & \Exp_S[|h(Z) - f_S(Z)|] + n_S], \\
    -\eps_T(h) \in & \; [-\Exp_T[|h(Z) - f_T(Z)|] - n_T, \\
    & -\Exp_T[|h(Z) - f_T(Z)|] + n_T]
\end{align*}
Hence,
\begin{align*}
     & \big|\eps_S(h)-\eps_T(h)\big| \leq |n_S + n_T| + \\
     & \big|\Exp_S[|h(Z) - f_S(Z)|]-\Exp_T[|h(Z) - f_T(Z)|]\big|
\end{align*}
Now to simplify the notation, for $e\in\{S, T\}$, define $\epsilon_e(h, h') = \Exp_e[|h(Z) - h'(Z)|]$, so that
\begin{align*}
    \big|\Exp_S[|h(Z) - f_S(Z)|]-\Exp_T[|h(Z) - f_T(Z)|]\big| \\
    = \big|\epsilon_S(h, f_S) - \epsilon_T(f_T, h)\big|.
\end{align*}
To bound $\big|\epsilon_S(h, f_S) - \epsilon_T(f_T, h)\big|$, on one hand, we have:
\begin{align*}
    & \big|\epsilon_S(h, f_S) - \epsilon_T(f_T, h)\big| = \\
    & \big|\epsilon_S(h, f_S) - \epsilon_S(h, f_T) + \epsilon_S(h, f_T) - \epsilon_T(f_T, h)\big| \\
&\leq \big|\epsilon_S(h, f_S) - \epsilon_S(h, f_T)\big| + \big|\epsilon_S(h, f_T) - \epsilon_T(f_T, h)\big| \\
&\leq \Exp_S[|f_S(Z) - f_T(Z)|] + \big|\epsilon_S(h, f_T) - \epsilon_T(f_T, h)\big| \\
\intertext{From \ref{lemma:lemma_a1}, we have:}
&\leq \Exp_S[|f_S(Z) - f_T(Z)|] +   d_{\HH \Delta \HH} (\mathcal{D}_S(Z), \mathcal{D}_T(Z)). \\
\end{align*}
Similarly, by the same trick of subtracting and adding back $\epsilon_T(h, f_S)$ above, the following inequality also holds:
\begin{align*}
    & \big|\epsilon_S(h, f_S) - \epsilon_T(f_T, h)\big| \leq \\
    & \Exp_T[|f_S(Z) - f_T(Z)|] +  d_{\HH \Delta \HH} (\mathcal{D}_S(Z), \mathcal{D}_T(Z)).
\end{align*}
Combine all the inequalities above, we know that:
\begin{align*}
    & \big|\eps_S(h)-\eps_T(h)\big| \leq |n_S + n_T| +  d_{\HH \Delta \HH} (\mathcal{D}_S(Z), \mathcal{D}_T(Z))  \\
&+ \min\{\Exp_S[|f_S(Z) - f_T(Z)|], \Exp_T[|f_S(Z) - f_T(Z)|]\}
\end{align*}
\end{proof}

\begin{lemma} 
\label{lemma:lemma_a3}
[\cite{mohri2018}, Corollary 3.19] Let $h \in \mathcal{H}:=\{h: \hdata \rightarrow\{0, 1\}\}$, where $VCdim(\HH) = d$. Then $\forall h \in \mathcal{H}, \forall \delta > 0$, w.p.b. at least $1 - \delta$ over the choice of a sample size $m$ and natural exponential $e$, the following inequality holds:
\begin{equation*}
\eps(h) \leq \widehat{\eps}(h) + \sqrt{\frac{2d}{m} \log \frac{em}{d}} + \sqrt{\frac{1}{2m}\log \frac{1}{\delta}}.
\end{equation*}
\end{lemma}

\begin{lemma}
Let $h \in \mathcal{H}:=\{h: \hdata \rightarrow\{0, 1\}\}$, where $VCdim(\HH) = d$. Then $\forall h \in \mathcal{H}, \forall \delta > 0$, then w.p.b. at least $1 - \delta$ over the choice of a sample size $m$ and natural exponential $e$, the following inequality holds:
\label{lemma:lemma_a4}
\begin{align*}
    \eps_T(h) \leq&~ \emperr_S(h) +  d_{\HH \Delta \HH} (\mathcal{D}_S(Z), \mathcal{D}_T(Z)) \\
    & + \min\{\Exp_S[|f_S(Z) - f_T(Z)|], \Exp_T[|f_S(Z) - f_T(Z)|]\} \\
    & + |n_S + n_T| + \sqrt{\frac{2d}{n} \log \frac{en}{d}} + \sqrt{\frac{1}{2n}\log \frac{1}{\delta}}
\end{align*}
\end{lemma}
\begin{proof}
Invoking the upper bound in \ref{lemma:lemma_a2}, we have w.p.b at least $1 - \delta$:
\begin{align*}
    \eps_T(h) \leq~& \eps_S(h) +  d_{\HH \Delta \HH} (\mathcal{D}_S(Z), \mathcal{D}_T(Z)) \\
    &+ \min\{\Exp_S[|f_S(Z) - f_T(Z)|], \Exp_T[|f_S(Z) - f_T(Z)|]\} \\
    &+ |n_S + n_T| \\
    \leq~& \emperr_S(h) +  d_{\HH \Delta \HH} (\mathcal{D}_S(Z), \mathcal{D}_T(Z)) \\
    &+ \min\{\Exp_S[|f_S(Z) - f_T(Z)|], \Exp_T[|f_S(Z) - f_T(Z)|]\} \\
    &+ |n_S + n_T| + \sqrt{\frac{2d}{n} \log \frac{en}{d}} + \sqrt{\frac{1}{2n}\log \frac{1}{\delta}}
\end{align*}
\end{proof}

\begin{theorem}
Let $h \in \mathcal{H}:=\{h: \hdata \rightarrow\{0, 1\}\}$, where $VCdim(\HH) = d$. For $0 < \delta < 1$, then w.p.b. at least $1 - \delta$ over the draw of samples $S$ and $T$, for all $h \in \HH$, we have:
 \label{thm:classification_upper_bound}
\begin{align*}
    \eps_T(h) \leq&~ \frac{m}{n+m}\emperr_T(h) + \frac{n}{n + m}\emperr_S(h) \\ 
    &+ \frac{n}{n + m} ( d_{\HH\Delta\HH} (\dist_S, \dist_T) \\
    &+ \min\{\Exp_S[|f_S(Z) - f_T(Z)|], \Exp_T[|f_S(Z) - f_T(Z)|]\} ) \\
    &+ \frac{n}{n+m} |n_S + n_T| \\
    &+ O\left(\sqrt{(\frac{1}{m} + \frac{1}{n}) \text{log} \frac{1}{\delta} + \frac{d}{n} \text{log}\frac{n}{d} + \frac{d}{m} \text{log}\frac{m}{d}} \right).
\end{align*}
\end{theorem}
\begin{proof}
\label{proof:thm_cls}
Having \ref{lemma:lemma_a3}, \ref{lemma:lemma_a4}, we can use a union bound to combine them with coefficients $m / (n + m)$ and $n / (n + m)$ respectively, we have:
\begin{align*}
    \eps_T(h) &~\leq \frac{m}{n+m} \left(\emperr_T(h) + \sqrt{\frac{2d}{m} \log \frac{em}{d}} + \sqrt{\frac{1}{2m}\log \frac{1}{\delta}}\right) \\
    & + \frac{n}{n + m} (\emperr_S(h) + d_{\HH\Delta\HH} (\dist_S, \dist_T) \\
    & + \min\{\Exp_S[|f_S(Z) - f_T(Z)|], \Exp_T[|f_S(Z) - f_T(Z)|]\}) \\ 
    &+ \frac{n}{n + m} \left(|n_S + n_T| + \sqrt{\frac{2d}{n} \log \frac{en}{d}} + \sqrt{\frac{1}{2n}\log \frac{1}{\delta}}\right). \\
\intertext{From Cauchy-Schwartz inequality, we obtain}
    \eps_T(h) &~\leq \frac{m}{n+m} \left(\emperr_T(h) + \sqrt{\frac{4d}{m} \log \frac{em}{d} + \frac{1}{m}\log \frac{1}{\delta}}\right) \\
    & + \frac{n}{n + m} (\emperr_S(h) + d_{\HH\Delta\HH} (\dist_S, \dist_T) \\
    & + \min\{\Exp_S[|f_S(Z) - f_T(Z)|], \Exp_T[|f_S(Z) - f_T(Z)|]\}) \\ 
    &+ \frac{n}{n + m} \left(|n_S + n_T| + \sqrt{\frac{4d}{n} \log \frac{en}{d} + \frac{1}{n}\log \frac{1}{\delta}}\right). \\
\intertext{As $m \ll n$ and applying Cauchy-Schwartz inequality one more time, we have}
    &~\leq \frac{m}{n+m}\emperr_T(h) + \frac{n}{n + m}\emperr_S(h) \\ 
    & + \frac{n}{n + m} (d_{\HH\Delta\HH} (\dist_S, \dist_T) + \\
    & \min\{\Exp_S[|f_S(Z) - f_T(Z)|], \Exp_T[|f_S(Z) - f_T(Z)|]\}) \\ 
    &+ \frac{n}{n + m} (|n_S + n_T| \\
    &+ \sqrt{\frac{8d}{m} \log \frac{em}{d} + \frac{2}{m}\log \frac{1}{\delta} + \frac{8d}{n} \log \frac{en}{d} + \frac{2}{n}\log \frac{1}{\delta}}). \\
    \leq&~ \frac{m}{n+m}\emperr_T(h) + \frac{n}{n + m}\emperr_S(h) \\ 
    & + \frac{n}{n + m} (d_{\HH\Delta\HH} (\dist_S, \dist_T) \\
    & + \min\{\Exp_S[|f_S(Z) - f_T(Z)|], \Exp_T[|f_S(Z) - f_T(Z)|]\}) \\ 
    & + \frac{n}{n + m} \left(|n_S + n_T|\right) \\
    & + O(\sqrt{(\frac{1}{m} + \frac{1}{n}) \text{log} \frac{1}{\delta} + \frac{d}{m} \text{log}\frac{m}{d} + \frac{d}{n} \text{log}\frac{n}{d}})
\end{align*}
\end{proof}

\subsection{Proof of regression bound}
For regression generalization bound, we follow the proof strategy in previous section, but with slight change of definitions. We let $\HH = \{h: \hdata \rightarrow [0, 1]\}$ be a set of bounded real-valued functions from the input space $\hdata$ to $[0, 1]$. We use $Pdim(\mathcal{H})$ to denote the pseudo-dimension of $\HH$, and let $Pdim(\mathcal{H}) = d$. We first prove the following lemmas that will be used in proving the main theorem:

\begin{restatable}{lemma}{first_lemma}
\label{lemma:lemma_a5}
\cite{zhao2018adversarial} For $h, h' \in \mathcal{H}:=\{h: \hdata \rightarrow[0, 1]\}$, where $Pdim(\mathcal{H}) = d$, and for any distribution $\mathcal{D}_S(Z)$, $\mathcal{D}_T(Z)$ over $\hdata$,
\begin{align*}
    |\epsilon_S(h, h') - \epsilon_T(h, h')| \leq  d_{\tilde{\mathcal{H}}} (\mathcal{D}_S(Z), \mathcal{D}_T(Z))
\end{align*}
where $\tilde{\mathcal{H}} := \{\mathbb{I}_{|h(x) - h'(x)| > t} : h, h' \in \mathcal{H}, 0 \leq t \leq 1 \}.$
\end{restatable}
\begin{lemma}
\label{lemma:lemma_a6}
For $h, h' \in \mathcal{H}:=\{h: \hdata \rightarrow[0, 1]\}$, where $Pdim(\mathcal{H}) = d$, and for any distribution $\mathcal{D}_S(Z)$, $\mathcal{D}_T(Z)$ over $\hdata$, we define $\tilde{\mathcal{H}} := \{ \mathbb{I}_{|h(x) - h'(x)| > t} := h, h' \in \HH, 0 \leq t \leq 1 \}$. Then $\forall h \in \mathcal{H}$, the following inequality holds:
\begin{align*}
& \big|\eps_S(h)-\eps_T(h)\big| \leq |n_S + n_T| +  d_{\tilde{\mathcal{H}}}  (\mathcal{D}_T(Z), \mathcal{D}_S(Z)) \\
&+ \min\{\Exp_S[|f_S(Z) - f_T(Z)|], \Exp_T[|f_S(Z) - f_T(Z)|]\}
\end{align*}
\end{lemma}

\begin{lemma} 
\label{lemma:lemma_a7}
Thm.11.8~\cite{mohri2018} Let $\HH$ be the set of real-valued function from $\hdata$ to $[0, 1]$. Assume that $Pdim(\HH) = d$. Then $\forall h \in \mathcal{H}, \forall \delta > 0$, with probability at least $1 - \delta$ over the choice of a sample size $m$ and natural exponential $e$, the following inequality holds:
\begin{equation*}
\eps(h) \leq \widehat{\eps}(h) + \sqrt{\frac{2d}{m} \log \frac{em}{d}} + \sqrt{\frac{1}{2m}\log \frac{1}{\delta}}.
\end{equation*}
\end{lemma}

\begin{lemma}
Let $\HH$ be a set of real-valued functions from $\hdata$ to $[0, 1]$ with $Pdim(\HH) = d$, and $\tilde{\HH} := \{\mathbb{I}_{|h(x) - h'(x)| > t} : h, h' \in \HH, 0 \leq t \leq 1 \}$. For $0 < \delta < 1$, then w.p.b. at least $1 - \delta$ over the draw of samples $S$ and $T$, for all $h \in \HH$, we have:
\label{lemma:lemma_a8}
\begin{align*}
    \eps_T(h) \leq&~ \emperr_S(h) + d_{\tilde{\HH}} (\dist_S, \dist_T) \\
    &+ \min\{\Exp_S[|f_S(Z) - f_T(Z)|], \Exp_T[|f_S(Z) - f_T(Z)|]\} \\
    &+ |n_S + n_T| + \sqrt{\frac{2d}{n} \log \frac{en}{d}} + \sqrt{\frac{1}{2n}\log \frac{1}{\delta}}
\end{align*}
\end{lemma}
\begin{proof}
Invoking the upper bound in \ref{lemma:lemma_a6} and \ref{lemma:lemma_a7}, we have w.p.b at least $1 - \delta$:
\begin{align*}
    \eps_T(h) \leq~& \emperr_S(h) + d_{\tilde{\HH}} (\dist_S, \dist_T) \\
    & + \min\{\Exp_S[|f_S(Z) - f_T(Z)|], \Exp_T[|f_S(Z) - f_T(Z)|]\} \\
    & + |n_S + n_T| \\
    \leq~& \emperr_S(h) + d_{\tilde{\HH}} (\dist_S, \dist_T) \\
    & + \min\{\Exp_S[|f_S(Z) - f_T(Z)|], \Exp_T[|f_S(Z) - f_T(Z)|]\} \\
    & + |n_S + n_T| + \sqrt{\frac{2d}{n} \log \frac{en}{d}} + \sqrt{\frac{1}{2n}\log \frac{1}{\delta}}
\end{align*}
\end{proof}
\begin{theorem}
Let $\HH$ be a set of real-valued functions from $\hdata$ to $[0, 1]$ with $Pdim(\HH) = d$, and $\tilde{\HH} := \{\mathbb{I}_{|h(x) - h'(x)| > t} : h, h' \in \HH, 0 \leq t \leq 1 \}$. For $0 < \delta < 1$, then w.p.b. at least $1 - \delta$ over the draw of samples $S$ and $T$, for all $h \in \HH$, we have:
 \label{thm:regression_upper_bound}
\begin{align*}
    \eps_T(h) \leq&~ \frac{m}{n+m}\emperr_T(h) + \frac{n}{n + m}\emperr_S(h) \\ 
    & + \frac{n}{n + m} (d_{\tilde{\HH}} (\dist_S, \dist_T) \\
    & + \min\{\Exp_S[|f_S(Z) - f_T(Z)|], \Exp_T[|f_S(Z) - f_T(Z)|]\}) \\
    & + \frac{n}{n+m} |n_S + n_T| + \\
    & O \left( \sqrt{(\frac{1}{m} + \frac{1}{n}) \text{log} \frac{1}{\delta} + \frac{d}{n} \text{log}\frac{n}{d} + \frac{d}{m} \text{log}\frac{m}{d}} \right).
\end{align*}
\end{theorem}
\begin{proof}
Having \ref{lemma:lemma_a5}, \ref{lemma:lemma_a6}, \ref{lemma:lemma_a7}, \ref{lemma:lemma_a8}, we can use a union bound to combine them with coefficients $m / (n + m)$ and $n / (n + m)$ respectively, and replace the $d_{\HH \Delta \HH} (\dist_S, \dist_T)$ with $d_{\tilde{\HH}} (\dist_S, \dist_T)$ in the proof of Thm.~\ref{proof:thm_cls}. Obviously, we have Thm.~\ref{thm:regression_upper_bound}. 
\end{proof}

\subsection{Discussions}
\paragraph{From binary to multi-class}
We leave the extension from binary to multi-class in the future work. Definition 4.1 is essentially a relaxed version of the classic total variation distance, so it can also be used in multi-class classification problems where we have more than 2 conditional distributions. However, the results in Theorem 4.1 indeed can only be applied in the binary classification setting, due to the use of VC-dimension, which only makes sense for binary classification problems. That being said, extension to the multi-class classification problem shouldn't be hard, although its presentation would be much more involved. At a high level, the proof essentially boils down to replace the VC dimension with the the Natarajan dimension~\cite{daniely2011multiclass}.

\paragraph{Generalizing to $\ell_1$ loss} The bound can indeed be straightforwardly generalized to $\ell_p$ loss. We use the $\ell_1$ loss in the presentation in order to be consistent with our experiments of vehicle counting, where we use the mean absolute error. A similar variational formulation of the conditional mutual information for continuous variables with $\ell_p$ loss also holds~\cite{farnia2016minimax}. We use the cross-entropy loss in practice for classification problems because the 0-1 loss is computationally intractable to optimize, and the cross-entropy loss serves as a convex proxy for the binary loss.

\paragraph{Bridging theory and practice} The bounds serve as justification to design our algorithms but they contain terms that are not directly available in practice, e.g., the optimal hypothesis of the two domains. On the other hand, using information-theoretic tools, we can close the gap between theory and practice, since the minimization of the conditional mutual information implies the minimization of the distance between the two optimal hypothesis.

\section{More Experimental Results}
\subsection{Comparison with State-of-the-art Methods}
Here we should address that we are not strictly in the same setting with MME~\cite{saito2019semi} and its related works~\cite{kim2020attract}. In all datasets, MME addresses in the few-shot setting and uses 1 or 3-shot as target labeled data. We argue that 1 or 3-shot is only suitable for few-shot learning and prototypical representations learning. The data selection will bias the model to a large extent. While we use more data to alleviate the selection bias and to mimic the usage in real domain adaptation application, such as 1\%, 5\% and more proportion of target labeled data. In DomainNet, MME use 126 out of the total 345 classes data, while we keep use the 345 classes for better evaluate our model performance in a more challenging setting. Although not strictly under the same setting, we also compare ours with the current state-of-the-art Semi-DA methods in our setting. MME~\cite{saito2019semi} and other related works (such as APE~\cite{kim2020attract}) assumes that there exists class-wise prototypical representations between source and target domain, and exploit Cosine Classifier~\cite{gidaris2018dynamic} to help learn the prototypical representations. However, our method can also seamlessly combine with Cosine Classifier for better performance, even comparing with the methods above as in Table~\ref{tab:sota_cls}.

\begin{table}[!htb]
\centering\footnotesize
\caption{The weights trade off between invariant representation part and invariant risk part under OfficeHome: Art to Real scenarios.(mean $\pm$ std)}
\begin{tabular}{c|ccc}
\toprule
& \multicolumn{3}{c}{$\lambda_{\text{risk}}$} \\ 
\cline{2-4} 
$\lambda_{\text{rep}}$ & 1 & 0.1 & 0.01  \\
\hline
1 & 70.23\scriptsize{$\pm$0.18} & 70.96\scriptsize{$\pm$0.17} & 70.55\scriptsize{$\pm$0.18} \\
0.1 & 71.20\scriptsize{$\pm$0.14} & 72.66\scriptsize{$\pm$0.16} & 72.31\scriptsize{$\pm$0.19} \\
0.01 & 72.65\scriptsize{$\pm$0.15} & \textbf{73.12}\scriptsize{$\pm$0.19} & 72.97\scriptsize{$\pm$0.20} \\
\hline
\end{tabular}
\label{tab:trade_off}
\end{table}

\begin{table*}[!htb]
\renewcommand\tabcolsep{4 pt} 
\centering\footnotesize
\caption{Accuracy (\%) comparison (higher means better) on \textbf{NICO}, \textbf{OfficeHome} with 1\% (above) and 5\% (below) labeled target data (mean $\pm$ std). Highest accuracies are highlighted in bold.}
\resizebox{\linewidth}{!}{%
\begin{tabular}{l|cc|cc|ccc}
\toprule
1\% \scriptsize{labeled target} & \multicolumn{2}{c|}{\centering NICO Animal}& \multicolumn{2}{c|}{\centering NICO Traffic} & \multicolumn{3}{c}{\centering OfficeHome} \\
\hline
 Method & \scriptsize{Grass to Snow} & \scriptsize{Snow to Grass} & \scriptsize{Sunset to Beach} & \scriptsize{Beach to Sunset} &\scriptsize{Art to Real} & \scriptsize{Real to Prod.} & \scriptsize{Prod. to Clip.} \\
 MME~\cite{saito2019semi} & 87.12\scriptsize{$\pm$0.76} & 79.52\scriptsize{$\pm$0.43} & 78.69\scriptsize{$\pm$0.86} & 74.21\scriptsize{$\pm$0.78} & 72.66\scriptsize{$\pm$0.18 }& 78.07\scriptsize{$\pm$0.17 }& 52.78\scriptsize{$\pm$0.16 } \\
APE~\cite{kim2020attract} & 87.41\scriptsize{$\pm$0.42} & 80.22\scriptsize{$\pm$0.45} & 79.70\scriptsize{$\pm$0.80} & 75.50\scriptsize{$\pm$0.85} & \textbf{74.41}\scriptsize{$\pm$0.49} & 78.45\scriptsize{$\pm$0.42} & \textbf{54.84}\scriptsize{$\pm$0.43} \\
\hline
Ours (LIRR + CosC) & \textbf{89.67}\scriptsize{$\pm$0.72} & \textbf{89.73}\scriptsize{$\pm$0.68} & \textbf{81.00}\scriptsize{$\pm$0.89} & \textbf{79.98}\scriptsize{$\pm$0.95} & 73.62\scriptsize{$\pm$0.21 }& \textbf{80.20}\scriptsize{$\pm$0.23 }& 53.84\scriptsize{$\pm$0.19 }  \\
\hline
\hline
\toprule
5\% \scriptsize{labeled target} & \multicolumn{2}{c|}{\centering NICO Animal}& \multicolumn{2}{c|}{\centering NICO Traffic} & \multicolumn{3}{c}{\centering OfficeHome} \\
\hline
 Method & \scriptsize{Grass to Snow} & \scriptsize{Snow to Grass} & \scriptsize{Sunset to Beach} & \scriptsize{Beach to Sunset} &\scriptsize{Art to Real} & \scriptsize{Real to Prod.} & \scriptsize{Prod. to Clip.} \\
\hline
MME~\cite{saito2019semi} & 87.80\scriptsize{$\pm$0.87} & 85.50\scriptsize{$\pm$0.95} & 92.02\scriptsize{$\pm$0.85} & 90.76\scriptsize{$\pm$0.81} & 75.24\scriptsize{$\pm$0.22} & 82.45\scriptsize{$\pm$0.18} & 61.75\scriptsize{$\pm$0.19} \\
APE~\cite{kim2020attract} & 87.85\scriptsize{$\pm$0.55} & 87.21\scriptsize{$\pm$0.80} & 92.55\scriptsize{$\pm$0.75} & 91.05\scriptsize{$\pm$0.45} & 75.90\scriptsize{$\pm$0.50} & \textbf{83.65}\scriptsize{$\pm$0.49} & 62.76\scriptsize{$\pm$0.51}\\
\hline
Ours (LIRR + CosC) & \textbf{88.97}\scriptsize{$\pm$0.45} & \textbf{88.22}\scriptsize{$\pm$0.55} & \textbf{92.70}\scriptsize{$\pm$0.87} & \textbf{91.50}\scriptsize{$\pm$1.05} & \textbf{76.63}\scriptsize{$\pm$0.19} & 83.45\scriptsize{$\pm$0.22} & \textbf{62.84}\scriptsize{$\pm$0.23} \\
\hline
\end{tabular}
}
\label{tab:sota_cls}
\end{table*}

\begin{table*}[!htb]
\renewcommand\tabcolsep{4 pt} 
\centering\footnotesize
\caption{Accuracy (\%) comparison (higher means better) on \textbf{DomainNet}, and \textbf{VisDA2017} with 1\% (above) and 5\% (below) labeled target data (mean $\pm$ std). Highest accuracies are highlighted in bold.}
\resizebox{0.8\linewidth}{!}{%
\begin{tabular}{l|ccc|c}
\toprule
1\% \scriptsize{labeled target} & \multicolumn{3}{c|}{\centering Domainnet} & \multicolumn{1}{c}{\centering VisDA2017} \\
\hline
 Method & \scriptsize{Real to Clip.} & \scriptsize{Sketch to Real} & \scriptsize{Clip. to Sketch}  & \scriptsize{Train to Val.} \\
 MME~\cite{saito2019semi} & 51.04\scriptsize{$\pm$0.12 }& 60.35\scriptsize{$\pm$0.12 }& 45.09\scriptsize{$\pm$0.14 } & 80.52\scriptsize{$\pm$0.35 } \\
APE~\cite{kim2020attract} & 52.21\scriptsize{$\pm$0.37} & \textbf{62.40}\scriptsize{$\pm$0.24} & 46.24\scriptsize{$\pm$0.42} & 81.09\scriptsize{$\pm$0.17} \\
\hline
Ours (LIRR + CosC) & \textbf{53.42}\scriptsize{$\pm$0.09 }& 61.79\scriptsize{$\pm$0.11 }& \textbf{47.83}\scriptsize{$\pm$0.10 } & \textbf{82.31}\scriptsize{$\pm$0.21 } \\
\hline
\hline
\toprule
5\% \scriptsize{labeled target} & \multicolumn{3}{c|}{\centering Domainnet} & \multicolumn{1}{c}{\centering VisDA2017} \\
\hline
 Method & \scriptsize{Real to Clip.} & \scriptsize{Sketch to Real} & \scriptsize{Clip. to Sketch}  & \scriptsize{Train to Val.} \\
\hline
MME~\cite{saito2019semi} & 62.31\scriptsize{$\pm$0.11} & 69.02\scriptsize{$\pm$0.18} & 53.88\scriptsize{$\pm$0.14} & 84.12\scriptsize{$\pm$0.22} \\
APE~\cite{kim2020attract} & 63.01\scriptsize{$\pm$0.23} & 68.92 \scriptsize{$\pm$0.22} & 53.95\scriptsize{$\pm$0.27} & 84.79\scriptsize{$\pm$0.49} \\
\hline
Ours (LIRR + CosC) & \textbf{63.03}\scriptsize{$\pm$0.17} & \textbf{69.52}\scriptsize{$\pm$0.09} & \textbf{54.44}\scriptsize{$\pm$0.12} & \textbf{85.06}\scriptsize{$\pm$0.17} \\
\hline
\end{tabular}
}
\label{tab:sota_cls_2}
\end{table*}

\subsection{Hyper-parameters}
There are two fundamental part in our proposed LIRR loss. One is the invariant representation item, the other is the invariant risk item. We use $\lambda_{\text{rep}}$ and $\lambda_{\text{risk}}$ represent the weights of invariant representation item and invariant risk item respectively. In order to explore the best trade off between this two items, we conduct extra experiments on Art to Real scenario in OfficeHome dataset. All other hyper-parameters settings are set as same as Sec.6.3. The results can be found in Table.~\ref{tab:trade_off}. From which, we can see that the optimal performance is achieved when $\lambda_{\text{risk}} = 0.1$ and $\lambda_{\text{rep}} = 0.01$.

\subsection{Implementation Details}
\textbf{For image classification task}: we use ResNet34 as backbone networks. We adopt SGD with learning rate of 1e-3, momentum of 0.9 and weight decay factor of 5e-4. We decay the learning rate with a multiplier 0.1 when training process reach three quarters of the total iterations. The batch size is set as 128 for VisDA2017 and Domainnet, 64 for officehome. For adversarial training, we use gradient reversal layer (GRL) to flip gradient in the backpropagation between feature encoder $g(\cdot)$ and domain discriminator $\mathcal{C}(\cdot)$ to obtain domain-invariant representation w.r.t. source labeled data and target unlabeled data. For min-max training objective for $\mathcal{L}_i$ and $\mathcal{L}_d$ in Eq.7, we implement it with the difference on two losses, $L(y, h(z))$ and $L(y, h(z, d))$. $h(z)$ is realized by a common predictor which only takes feature $z$ as input. $h(z, d)$ indicates an additional predictor which takes the combination of feature $z$ and domain index $d$, e.g. we concatenate original feature $z$ with an additional full $0$ (or $1$) channel to represent source(or target) domain. It's worth noting that according to ~\cite{saito2019semi}, the utilization of entropy minimization hurts the performance. Thus, we implement the CDAN method without entropy minimization. Our results are all obtained without heavy engineering tricks. All code is implemented in Pytorch and will be made available upon acceptance.

\textbf{For traffic counting regression task}: we use VGG16 as encoder and FCN8s~\cite{shelhamer2017fully} as decoder. The model will output a density map as the regression result for input images. The optimizing goal is a joint loss including both the euclidean loss between the groundtruth density map and the predicted one, and the mean absolute counting error loss between the total predicted count and groundtruth count. We use mean absolute error (MAE) metric for evaluation, which measure the absolute difference between the output count and the ground-truth count. We adopt Adam optimizer with learning rate set to 1e-6. The batch size is set as 24.

\subsection{Discussions}
\paragraph{LIRR with cosine classifier}
It should be noted that LIRR alone achieves favorable performance over other baselines. With the cosine classifier, a useful technique adopted by many previous works~\cite{saito2019semi}, LIRR's performance can be further improved. We perform this additional experiment mainly to show that our framework is also compatible with existing techniques, e.g., the cosine classifier.

\paragraph{Grad-CAM results}
In Figure.4, LIRR captures a more complete representation of the husky's face than both DANN and Source+Target, in which they capture only parts of the husky's face. Similarly for puppy (second row), LIRR captures more on the puppy's outline and body shape.

\end{document}